\documentclass[10pt,twocolumn,letterpaper]{article}

\usepackage{cvpr}              

\usepackage{graphicx}
\usepackage{amsmath}
\usepackage{amssymb}
\usepackage{booktabs}
\usepackage{placeins}

\renewcommand{\paragraph}[1]{{\vspace{1mm}\noindent\bf#1 \quad}}

\usepackage[pagebackref,breaklinks,colorlinks]{hyperref}

\usepackage[accsupp]{axessibility}

\usepackage[capitalize]{cleveref}
\crefname{section}{Sec.}{Secs.}
\Crefname{section}{Section}{Sections}
\Crefname{table}{Table}{Tables}
\crefname{table}{Tab.}{Tabs.}

\def\cvprPaperID{*****} 
\def\confName{CVPR}
\def\confYear{2022}

\begin{document}

\title{Neural Face Video Compression using Multiple Views}

\author{\hspace{-0.3cm}Anna Volokitin$^{*,1}$\!, Stefan Brugger$^2$\!, Ali Benlalah$^2$\!, Sebastian Martin$^2$\!, Brian Amberg$^2$\!, Michael Tschannen$^2$ \\
\hspace{-0.3cm}$^1$ETH Zurich, {\tt\small anna@volokitin.net} \, $^2$Apple, {\tt\small \{sbrugger,abenlalah,sebi,bamberg,mtschannen\}@apple.com}
}

\maketitle

\let\thefootnote\relax\footnotetext{\vspace{-0.2cm}$^*$Work done at Apple.}
\begin{abstract}
Recent advances in deep generative models led to the development of neural face video compression codecs that use an order of magnitude less bandwidth than engineered codecs. These neural codecs reconstruct the current frame by warping a source frame and using a generative model to compensate for imperfections in the warped source frame. Thereby, the warp is encoded and transmitted using a small number of keypoints rather than a dense flow field, which leads to massive savings compared to traditional codecs. However, by relying on a single source frame only, these methods lead to inaccurate reconstructions (e.g. one side of the head becomes unoccluded when turning the head and has to be synthesized). Here, we aim to tackle this issue by relying on multiple source frames (views of the face) and present encouraging results.
\end{abstract}

\section{Introduction}

Neural image and video compression research has made great strides over the past few years. In particular, the latest models leverage advances in deep generative modeling to produce neural compression models, which outperform their state-of-the-art engineered counterparts by significant margins in rate-distortion performance when taking human perceptual quality into account \cite{mentzer2020high, mentzer2021towards}. Even larger gains can be achieved by learning a domain-specific compression model when the data domain is constrained.

Designing and training compression models specific to video calls is one of the most recent breakthrough stories along these lines \cite{wang2021one,oquab2021low,konuko2021ultra}, with some works reporting an order of magnitude of bit rate reduction at a given perceptual quality compared to engineered codecs \cite{wang2021one}. In a nutshell, these face video compression algorithms rely on a source frame (view) of the face, warp this view to approximate the target frame to be transmitted, and process the warped source view (or features extracted from the source view) with a generator to compensate for imperfections in the warped source view. Thereby, the warp is parametrized using a small set of keypoints or local affine transforms, extracted from the target frame using a corresponding learned predictor, which enables highly efficient coding and transmission of the warp.

While these models achieve excellent (perceptual) rate-distortion performance in the average case, they have some clear limitations due to relying on single source view. Indeed, a single view might not provide appearance details about one side of the face if the head in the source view is slightly rotated to the left or right. Similarly, a source view showing the face with a closed mouth does not provide any information about the appearance of the mouth interior. More generally, it is usually impossible to faithfully infer person-specific shape and appearance details for different facial expressions by solely relying on a single view of the face \cite{zollhofer2018state}.

In this paper, we take a first step towards leveraging multiple views of the face to improve neural face video compression algorithms. Specifically, we explore two key questions of multi-view neural face compression, namely how to select different views, and how to fuse information across different views. We present ablations along these two axes, and show that our best 3-view model, which is based on a simple and computationally inexpensive backbone, outperforms the state-of-the art single frame model at significantly lower per-frame rate. Furthermore, we outline and discuss future research directions to push neural face compression algorithms to production quality.

\begin{figure}[t]
\vspace{-2mm}
  \centering
\includegraphics[width=0.6\columnwidth]{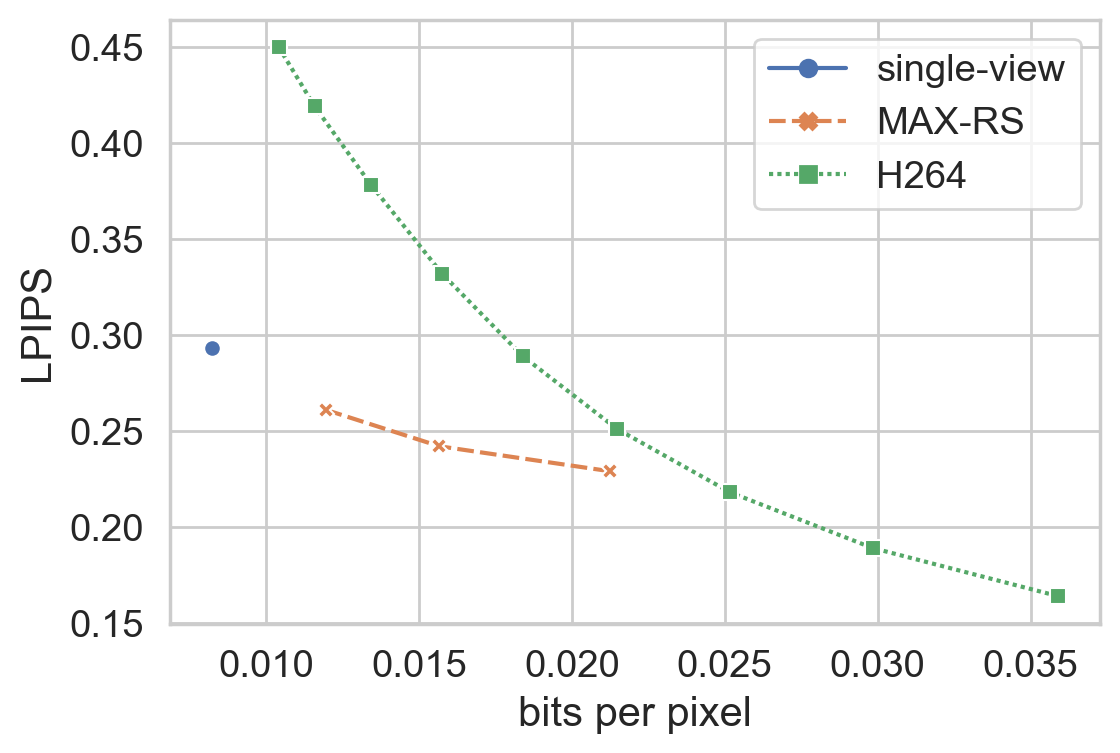}
\vspace{-2mm}
\caption[rd-curve]{Rate-distortion performance comparison of our single-frame baseline, our max pooling model based random view sampling (for 3, 5, and 8 source views), and H264.}
\label{fig:rd-curve}
\vspace{-4mm}
\end{figure}

\section{Related work}

While research on general neural video compression already features a rich body of literature (e.g. \cite{wu2018video,lu2019dvc,habibian2019video,rippel2021elfvc,mentzer2021towards}), there is only a handful of works on neural face video compression \cite{oquab2021low,konuko2021ultra,wang2021one}. Oquab \emph{et al.} \cite{oquab2021low} study the suitability of different talking head synthesis approaches for compression, targeting a mobile low-resource scenario. Konuko \emph{et al.} \cite{konuko2021ultra} propose a simple procedure to select one out of multiple source views used to reconstruct the target frame using exhaustive search on the reconstruction error at encoding time. Wang \emph{et al.} \cite{wang2021one} introduce a high-fidelity model which allows adapting the head pose while decoding.

Closely related to neural face video compression is (2D) talking head synthesis, which uses different mechanisms for motion transfer from a source to a target view, including warping, supervised or unsupervised keypoints, conditional normalization or a combination of these to \cite{pumarola2018ganimation, wiles2018x2face, zakharov2019few, wang2019few, siarohin2019first, zakharov2020fast, ha2020marionette}. The resulting motion representations can have various forms and are often not suited for compression. One line of work attempts to overcome inaccurate modeling of subject identity by relying on multiple frames \cite{wang2019few,ha2020marionette}, but no approach considers careful frame selection beyond random sampling.

\begin{figure*}[ht!]
\vspace{-0.1cm}
  \centering
\includegraphics[width=0.9\textwidth]{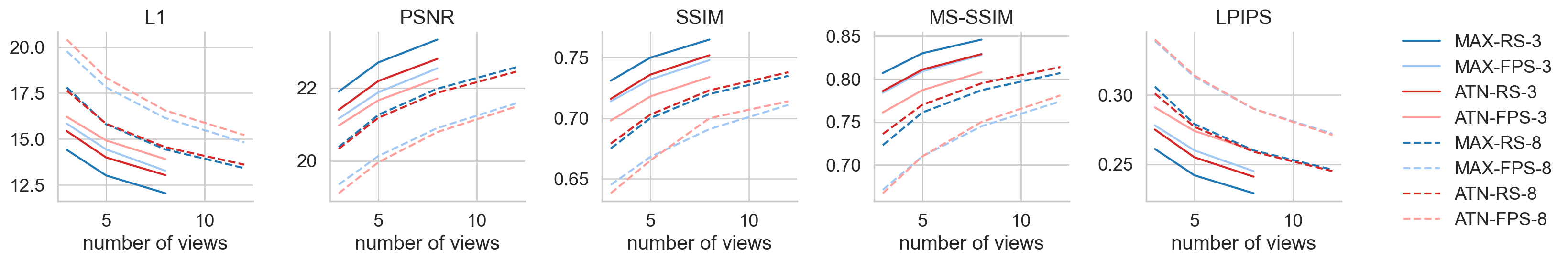}
\vspace{-0.3cm}
\caption[ablation]{How well do models trained with a fixed number of source views generalize to a different number of source views? This figure shows the performance of aggregation modules, source view sampling strategies, and number of training source views (legend format {\tt <aggregation strategy>-<sampling strategy>-<number of training source views>}) as a function of the number of testing source views. We report the L1 reconstruction error, PSNR, SSIM and MS-SSIM \cite{wang2003multiscale}, as well as LPIPS \cite{zhang2018unreasonable}. \vspace{-0.2cm}
}
\vspace{-0.2cm}
\label{fig:ablation}
\end{figure*}

\section{Method}
We build on top of the first order motion model (FOMM) from \cite{siarohin2019first}, which also underlies \cite{wang2021one}. We start by briefly summarizing the FOMM (cf. \cite[Fig.~2]{siarohin2019first}), which consists of a keypoint detector, a dense motion predictor, a generator and a corresponding encoder. Given a target frame $f$ and a source frame (view) $v$, the FOMM first extracts sparse keypoints and corresponding local affine transformations from $f$ and $v$, which together define a coarse flow field. The keypoints and transformations are then processed with the dense motion predictor, which refines the coarse flow field and also predicts an occlusion map. Finally, the flow is used to warp features extracted from $v$, mask them using the occlusion map, and to reconstruct $\hat f$ with the generator. Note that the keypoint extractor and the encoder only have to be applied once to $v$. In a video call, the source view is transmitted first, and all subsequent frames are encoded and transmitted via their corresponding keypoints and affines. The transmission cost of the source view is amortized across the entire call.

\begin{figure*}[ht!]
  \centering
\includegraphics[width=0.38\textwidth]{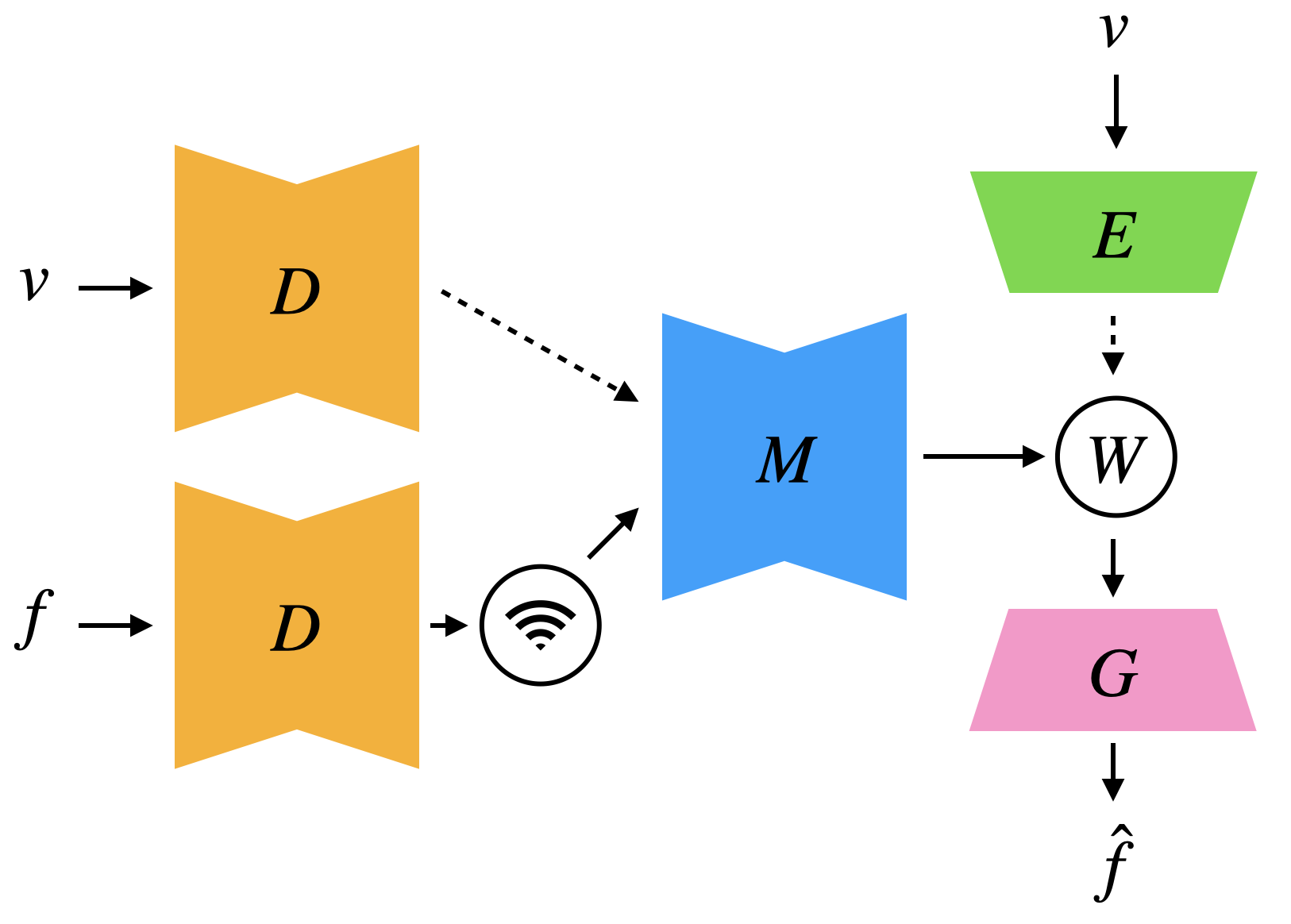}
\hspace{0.04\textwidth}
\includegraphics[width=0.57\textwidth]{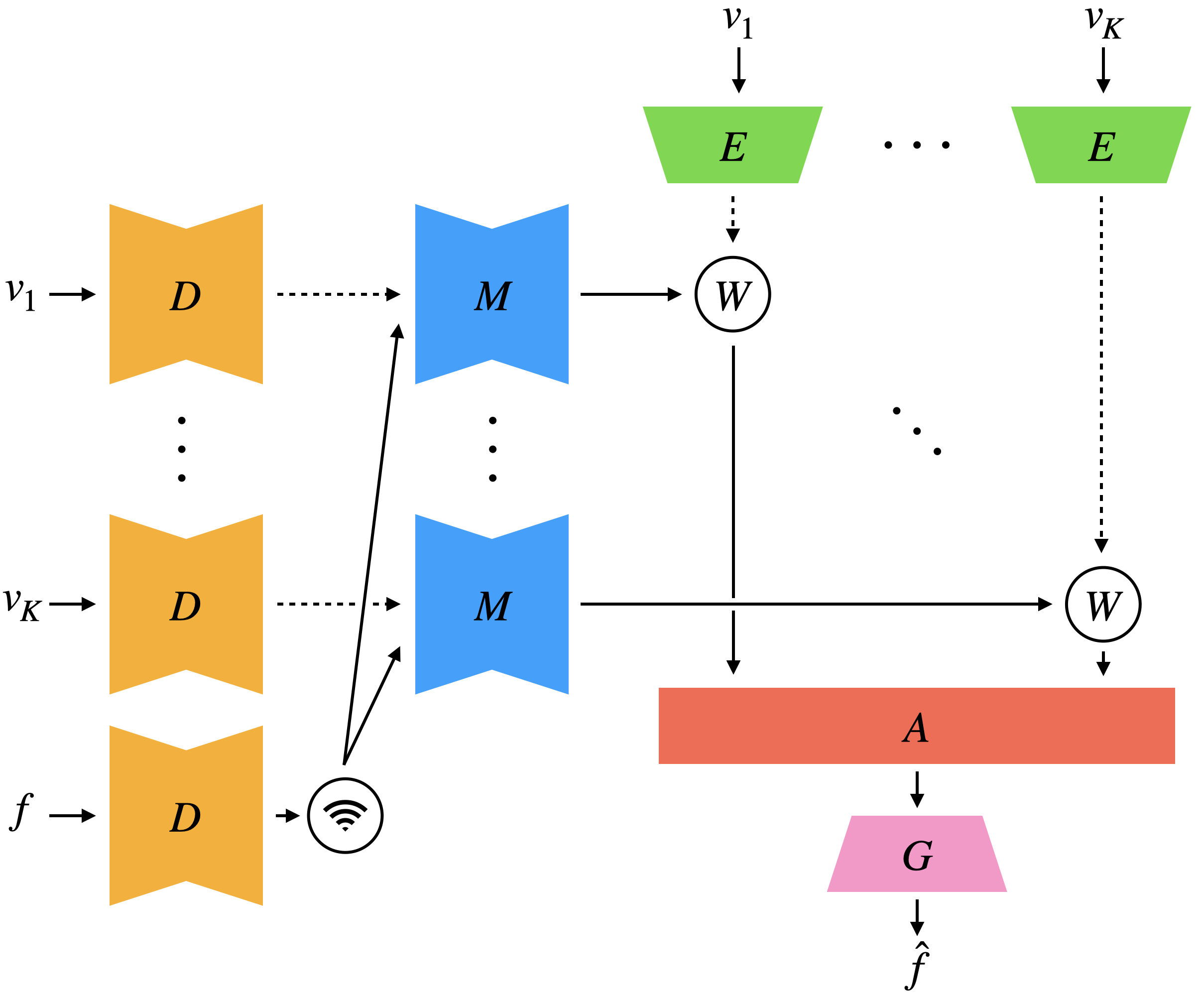}
\vspace{0.2cm}
\caption[schematic]{Schematic of the single-frame (left) and multi-frame (right) networks. The single-frame model is a precise re-implementation of the FOMM from \cite{siarohin2019first}. Both networks comprise a keypoint detector ($D$, UNet-based), a dense motion predictor ($M$, UNet-based), an encoder ($E$, ResNet-based), a corresponding generator ($G$, ResNet-based), as well as a warp-and-occlusion module ($W$). The multi-frame network additionally includes a view aggregation module ($A$, see Sec.~\ref{sec:aggregation}), with pooling or attention based aggregation mechanism. All blocks with the same color are shared. The quantities denoted by dashed arrows only have to be computed once per video, and the corresponding inputs (i.e., the source views $v$ and $v_1, \ldots, v_K$) hence only have to be transmitted once. For both single- and multi-frame models, the only quantity transmitted per frame to reconstruct the target frame $f$ are the corresponding keypoints (marked with a ``WiFi`` symbol).}
\label{fig:schematic}
\end{figure*}

To extend the FOMM to multiple source views $\{v_k\}_{k=1}^K$ we apply the keypoint detector to every view $v_k$ to obtain a coarse flow for every $k$ and corresponding refined flow by applying the dense motion predictor to every coarse flow. We then obtain warped and masked features for every view and pass them to an aggregation module (discussed in more detail below) whose output is fed to the generator to reconstruct the target frame $f$ (see Fig.~\ref{fig:schematic} for an overview).

Preliminary experiments showed that when using multiple source views, predicting local affine transformations does not improve performance, so we rely on keypoints only. 
Note that as for the single frame model we transmit only the keypoints for the current target frame; each view only has to be transmitted once (and corresponding keypoints only have to be extracted once) so that transmission costs are again amortized for reasonably long video sequences. The only additional cost we incur compared to the single frame model is computational, namely we need to compute dense flow fields occlusion maps for $K$ source views instead of one, and apply the aggregation module to fuse the warped features from the $K$ source views.

Given the architecture described in the preceding paragraph, the following two questions arise naturally.
\begin{itemize}
\item What is an effective and efficient architecture design for the aggregation module?
\item How to select the source views in order to achieve substantial improvements over just using a single source view?
\end{itemize}

In this paper, we take a first step in exploring these questions, as described next.

\subsection{View aggregation module \label{sec:aggregation}} The view aggregation module should have two properties, namely 1) it should be invariant to the order in which the views are aggregated, and 2) it should be able to aggregate a changing number of views as more views will arrive over time. Indeed, as discussed below, we are interested in having diverse views with complementary information, which will only become available over time, as the caller moves their head and shows different facial expressions.

We consider two different architectures. The first one amounts to simple, permutation-invariant pooling such as max pooling and average pooling \cite{zaheer2017deep}. In order to facilitate fusion of the features, we first apply the same stack of two residual blocks to the features extracted from every view after warping/masking, and pool the features across views.

\newcommand{\R}{\mathbb R}

The second pooling architecture we consider is based on self-attention (SA) as formulated in \cite{wang2018non}  between identical spatial locations in feature space, across views. This is in contrast to the more common SA across different spatial locations within one image. The motivation for attending to the same location across views is to allow our model to pixel-wise attend to the most useful (pre-warped) source view for reconstructing the target frame. In more detail, let $z_{i,j}^{(k)} \in \R^c$ be the feature extracted from view $k$ after warping and masking at spatial location $i,j$, and $Z_{i,j} = [z_{i,j}^{(1)}, \dots, z_{i,j}^{(K)}]^\top \in \R^{K \times c}$ the matrix obtained by stacking the $z_{i,j}^{(k)}$. We obtain query, key and value matrices as $Q_{i,j} = Z_{i,j}  W_q$, $K_{i,j} = Z_{i,j} W_k$, and $V_{i,j} = Z_{i,j} W_v$, respectively, where $W_q, W_k \in \R^{c \times d}, W_v \in \R^{c \times c}$. The SA output at spatial location $i,j$ in feature space, for all $K$ views, is then computed as
\begin{equation}
A_{i,j} = \text{softmax}(Q_{i,j} K_{i,j}^\top / \sqrt{d}) V_{i,j}.
\end{equation}
Note that computing $Q_{i,j}$, $K_{i,j}$, and $V_{i,j}$ over all spatial locations amounts to $1 \times 1$ convolution. To build a SA block, we add a residual branch to the SA output, apply ChannelNorm \cite{mentzer2020high}, and append another $3 \times 3$ convolution layer (maintaining the number of channels) followed by ChannelNorm. We stack two such blocks and apply a convolution producing a $W \times H \times K$ output, which we normalize along the channel dimension using softmax. Finally, we average the $z_{i,j}^{(k)}$ across views with the softmax output as weights.

\subsection{View selection strategies \label{sec:sampling}} We consider two different sampling strategies: Random sampling (RS) and furthest point sampling (FPS) based on facial landmarks. RS amounts to picking the first frame as well as $K-1$ frames at random among the remaining frames in the video sequence. 
For FPS \cite{katsavounidis1994new} we first extract 2D facial landmarks from each frame using a simple and efficient landmark detector similar to \cite{trigeorgis2016mnemonic}, and use the stacked coordinates of the landmarks as feature vectors for each frame. We then sample frames, selecting first the initial frame, and then at each iteration selecting the frame whose feature vector maximizes the minimal Euclidean distance to the feature vectors of the previously selected frames.

For this initial study we focus on batch (per-video) sampling algorithms for simplicity, and note that online algorithms are necessary to enable a real-time video call scenario. Online uniform RS can be realized using reservoir sampling \cite{li1994reservoir}; a similar reservoir-based streaming algorithm can be derived for FPS. Note that these online approaches likely require careful tuning to ensure a small enough (amortized) bit rate increase.

\section{Experiments}

\paragraph{Experimental setup} We use the VoxCeleb2 data set \cite{chung2018voxceleb2} which contains 150k videos of 6k celebrities, split into 1.1M individual speech sequences, cropped and scaled to $224 \times 224$ pixels, and split into a train and test set with disjoint subject identities. As we are interested in modeling real-world scenarios where video calls can span several minutes to hours, unless stated otherwise, we concatenate speech sequences of every video and use the resulting videos for our evaluations. These videos are generally harder to compress than individual speech sequences as the target frames may deviate more from a given set of source views as the video length increases.

Our baseline model is a precise re-implementation of the FOMM from \cite{siarohin2019first} without local affine transformations (as these did not significantly reduce reconstruction errors in our experiments). To obtain multi-frame models, we extend the baseline model with one of the view aggregation modules described in Sec.~\ref{sec:aggregation} and select views according to the strategies described in Sec.~\ref{sec:sampling}. We employ the multi-scale VGG-based perceptual loss from \cite{johnson2016perceptual} (following the implementation of \cite{siarohin2019first}) and the Adam optimizer, and select for every model the learning rate from $\{3\cdot10^{-6}, 1\cdot10^{-5},3\cdot10^{-5}\}$ that leads to the lowest training loss. We train our models on 4 V100 GPUs for 300k iterations with a batch size of 48 (we increase the number of iterations for smaller batch sizes such that the total number of training frames seen by all models is identical).

\paragraph{Comparison of aggregation / view selection strategies} We combine the max pooling and SA view aggregation modules each with RS and FPS, and train the so-obtained models using 3 or 8 source views. We then evaluate these models with a number of source views in the range $\{3, \ldots, 12\}$. The results are shown in Fig.~\ref{fig:ablation}. It can be seen that, while all models obtain smaller reconstruction errors as the number of source views increases at evaluation time, the model MAX-RS-3 (i.e. the model with view aggregation based on max pooling combined with RS, trained on 3 source views) performs best.

Upon visually inspecting the models that were trained with 8 source views, we observe that these models do not generalize well in situations where the target frame semantics differ significantly from those in the source views. Thus, it seems that the 8-view models do not see a diverse enough set of source views during training to learn to interpolate/aggregate the source views when the target frame semantics differ significantly from those in the source views. In contrast, models trained with 3 source views are forced to predict larger semantic changes during training.

One possible explanation for RS performing better than facial landmark-based FPS could be that while FPS is beneficial for reconstruction the face region, it might lead to an increase in reconstruction error in the background region, and thereby of the overall reconstruction error. As for the simple max pooling outperforming the more complex SA-based aggregation, we speculate that SA might lead to larger benefits in scenarios where the source views are not pre-aligned with the keypoint-based warp so that the SA module is applied to the unmodified source view features as in \cite{ha2020marionette}, handling both feature alignment and aggregation.

\paragraph{Comparison with prior work}
We compare our single-view baseline model and our best 3-view model (MAX-RS-3) with \cite{wang2021one, zakharov2020fast, siarohin2019first, wang2019few} in Table~\ref{tab:rel-work} when evaluated on single speech sequences rather than concatenated speech sequences (as this is the evaluation protocol in the referenced papers). \emph{Our MAX-RS-3 outperforms the closest competitor \cite{wang2021one} in three out of four metrics while using one third of the (uncoded) bits to encode the keypoints (our model uses 10 2D keypoints resulting in 20 floats per frame while \cite{wang2021one} uses 20 3D keypoints resulting in 60 floats) and arguably a simpler model}. Note however, that the model from \cite{wang2021one} allows re-rendering the head from a different (angular) viewpoint, which our model cannot do. Further \cite{wang2021one} attain substantial reduction in the bits transmitted per frame by making the number of keypoints used adaptive and by using entropy coding. Both techniques can be applied to our model as well and are expected to achieve similar reductions.

\paragraph{Comparison with H264} 
Work \cite{wang2021one} reports an order of magnitude rate reduction of their model over H264 at the same reconstruction quality in terms of learned perceptual image patch similarity (LPIPS) \cite{zhang2018unreasonable}, when trained and evaluated on a proprietary data set. Here, we compare the rate-distortion performance of our single view baseline as well as our best multi-view models MAX-RS-3/5/8 to H264 in Fig.~\ref{fig:rd-curve}. All of our models transmit the same information on a per-frame basis, namely 10 2D keypoints encoded as 16 bit floats. The only rate overhead incurred is due to the transmission of multiple source views. This rate overhead, even when amortized over hundreds of frames, still affects the rate distortion tradeoff quite significantly. However, asymptotically this overhead vanishes leaving a pure distortion complexity tradeoff, i.e., the blue marker and the orange curve would align vertically in Fig.~\ref{fig:rd-curve}. Further, our method obtains a rate reduction of only about $2\times$ over H264 which is a substantially smaller reduction than what is reported by \cite{wang2021one}. We reiterate that \cite{wang2021one} uses an adaptive number of keypoints as well as entropy coding, and evaluates on a higher-resolution data set.

\begin{table}[t]
    \centering
    \small
    \begin{tabular}{lcccc}
\toprule
 & L1 & PSNR & SSIM & MS-SSIM \\
 \midrule
fs-vid2vid \cite{wang2019few} & 17.10 & 20.36 & 0.710 & -- \\
FOMM \cite{siarohin2019first} & 12.66 & 23.25 & 0.770 & 0.830 \\
Bi-Layer \cite{zakharov2020fast} & 23.95 & 16.98 & 0.660 & 0.660 \\
NVIDIA \cite{wang2021one} & 10.74 & 24.37 & {\bf 0.800} & 0.850 \\
\midrule
single-frame & 13.16 & 22.79 & 0.758 & 0.828 \\
MAX-RS-3 & {\bf 10.00} & {\bf 24.87} & 0.795 & {\bf 0.872} \\
\bottomrule
\end{tabular}
\vspace{-1mm}
    \caption[tab-rel-work]{ \label{tab:rel-work}Comparison of our single-view baseline and best performing 3 view model with prior work on VoxCeleb2; the prior work values are taken from \cite[Table 1]{wang2021one}.}
\vspace{-4mm}
\end{table}

\section{Conclusions}

We presented a first exploration and evaluation of different approaches to selecting and aggregating information across views in the context of neural face video compression. Our best 3-view model outperforms the state-of-the-art single view model at a significantly lower per-frame bit rate. Furthermore, we identified multi-view models as an effective means to navigate the tradeoff between distortion and decoding complexity. We conclude by outlining future research directions.

\paragraph{Computationally efficient models} Neural models have excellent rate-distortion performance but still consume orders of magnitude more power than engineered codecs with dedicated hardware accelerators. Particularly important in the context of real-time on-device compression scenarios, such as video calls, is therefore to substantially improve the computational efficiency of neural compression models. We believe that relying on multiple views can help along these lines as less of the face modeling work needs to be done by the decoder model.

\paragraph{Online view selection} Developing more sophisticated view selection mechanisms is important to gradually gather fine-grained appearance information about the caller in order to produce faithful reconstructions reflecting person-specific expression details. This selection mechanism should be computationally inexpensive (it could e.g. rely on the learned keypoints underlying the warp) and should potentially allow updating views, e.g. when the background or the lighting situation changes. Online updates of source views will, however, incur a transmission overhead so they cannot occur often. As a consequence the neural codec will need to model uncertainty about the face regions or expressions it has not seen and reliably identify frames that provide additional information. Work \cite{konuko2021ultra} proposes an online selection mechanism, but incurs a large increase in encoding complexity and is not suited for real-time scenarios.

\paragraph{Better data sets} Whereas celebrity interview videos as abundantly available on the web and collected in the VoxCeleb2 data set present a good starting point, they are certainly not representative of all video call scenarios. To develop neural face codecs tailored to everyday situations data sets, featuring a broader set of video call situations (including e.g. calls from handheld devices while walking, and recorded with a large range of different cameras) are necessary. A corresponding public data set could greatly accelerate development of robust and product grade algorithms.

{\small
\bibliographystyle{ieee_fullname}
\bibliography{egbib}

\begin{thebibliography}{10}\itemsep=-1pt

\bibitem{chung2018voxceleb2}
Joon~Son Chung, Arsha Nagrani, and Andrew Zisserman.
\newblock Voxceleb2: Deep speaker recognition.
\newblock In {\em Interspeech}, 2018.

\bibitem{ha2020marionette}
Sungjoo Ha, Martin Kersner, Beomsu Kim, Seokjun Seo, and Dongyoung Kim.
\newblock Marionette: Few-shot face reenactment preserving identity of unseen
  targets.
\newblock In {\em AAAI Conference on Artificial Intelligence (AAAI)}, pages
  10893--10900, 2020.

\bibitem{habibian2019video}
Amirhossein Habibian, Ties~van Rozendaal, Jakub~M Tomczak, and Taco~S Cohen.
\newblock Video compression with rate-distortion autoencoders.
\newblock In {\em International Conference on Computer Vision (ICCV)}, pages
  7033--7042, 2019.

\bibitem{johnson2016perceptual}
Justin Johnson, Alexandre Alahi, and Li Fei-Fei.
\newblock Perceptual losses for real-time style transfer and super-resolution.
\newblock In {\em European Conference on Computer Vision (ECCV)}, pages
  694--711, 2016.

\bibitem{katsavounidis1994new}
Ioannis Katsavounidis, C-C~Jay Kuo, and Zhen Zhang.
\newblock A new initialization technique for generalized lloyd iteration.
\newblock {\em IEEE Signal Processing Letters}, 1(10):144--146, 1994.

\bibitem{konuko2021ultra}
Goluck Konuko, Giuseppe Valenzise, and St{\'e}phane Lathuili{\`e}re.
\newblock Ultra-low bitrate video conferencing using deep image animation.
\newblock In {\em IEEE International Conference on Acoustics, Speech and Signal
  Processing (ICASSP)}, pages 4210--4214, 2021.

\bibitem{li1994reservoir}
Kim-Hung Li.
\newblock Reservoir-sampling algorithms of time complexity {O}(n(1+log (n/n))).
\newblock {\em ACM Transactions on Mathematical Software (TOMS)},
  20(4):481--493, 1994.

\bibitem{lu2019dvc}
Guo Lu, Wanli Ouyang, Dong Xu, Xiaoyun Zhang, Chunlei Cai, and Zhiyong Gao.
\newblock Dvc: An end-to-end deep video compression framework.
\newblock In {\em IEEE Conference on Computer Vision and Pattern Recognition
  (CVPR)}, pages 11006--11015, 2019.

\bibitem{mentzer2021towards}
Fabian Mentzer, Eirikur Agustsson, Johannes Ball{\'e}, David Minnen, Nick
  Johnston, and George Toderici.
\newblock Towards generative video compression.
\newblock {\em arXiv:2107.12038}, 2021.

\bibitem{mentzer2020high}
Fabian Mentzer, George Toderici, Michael Tschannen, and Eirikur Agustsson.
\newblock High-fidelity generative image compression.
\newblock In {\em Neural Information Processing Systems (NeurIPS)}, 2020.

\bibitem{oquab2021low}
Maxime Oquab, Pierre Stock, Daniel Haziza, Tao Xu, Peizhao Zhang, Onur Celebi,
  Yana Hasson, Patrick Labatut, Bobo Bose-Kolanu, Thibault Peyronel, et~al.
\newblock Low bandwidth video-chat compression using deep generative models.
\newblock In {\em IEEE Conference on Computer Vision and Pattern Recognition
  Workshops (CVPRW)}, pages 2388--2397, 2021.

\bibitem{pumarola2018ganimation}
Albert Pumarola, Antonio Agudo, Aleix~M Martinez, Alberto Sanfeliu, and
  Francesc Moreno-Noguer.
\newblock Ganimation: Anatomically-aware facial animation from a single image.
\newblock In {\em European Conference on Computer Vision (ECCV)}, pages
  818--833, 2018.

\bibitem{rippel2021elfvc}
Oren Rippel, Alexander~G. Anderson, Kedar Tatwawadi, Sanjay Nair, Craig Lytle,
  and Lubomir Bourdev.
\newblock Elf-vc: Efficient learned flexible-rate video coding.
\newblock In {\em IEEE International Conference on Computer Vision (ICCV)},
  pages 14479--14488, October 2021.

\bibitem{siarohin2019first}
Aliaksandr Siarohin, St{\'e}phane Lathuili{\`e}re, Sergey Tulyakov, Elisa
  Ricci, and Nicu Sebe.
\newblock First order motion model for image animation.
\newblock In {\em Neural Information Processing Systems (NeurIPS)}, pages
  7137--7147, 2019.

\bibitem{trigeorgis2016mnemonic}
George Trigeorgis, Patrick Snape, Mihalis~A Nicolaou, Epameinondas Antonakos,
  and Stefanos Zafeiriou.
\newblock Mnemonic descent method: A recurrent process applied for end-to-end
  face alignment.
\newblock In {\em IEEE Conference on Computer Vision and Pattern Recognition
  (CVPR)}, pages 4177--4187, 2016.

\bibitem{wang2019few}
Ting-Chun Wang, Ming-Yu Liu, Andrew Tao, Guilin Liu, Bryan Catanzaro, and Jan
  Kautz.
\newblock Few-shot video-to-video synthesis.
\newblock In {\em Neural Information Processing Systems (NeurIPS)}, pages
  5013--5024, 2019.

\bibitem{wang2021one}
Ting-Chun Wang, Arun Mallya, and Ming-Yu Liu.
\newblock One-shot free-view neural talking-head synthesis for video
  conferencing.
\newblock In {\em IEEE Conference on Computer Vision and Pattern Recognition
  (CVPR)}, pages 10039--10049, 2021.

\bibitem{wang2018non}
Xiaolong Wang, Ross Girshick, Abhinav Gupta, and Kaiming He.
\newblock Non-local neural networks.
\newblock In {\em IEEE Conference on Computer Vision and Pattern Recognition
  (CVPR)}, pages 7794--7803, 2018.

\bibitem{wang2003multiscale}
Zhou Wang, Eero~P Simoncelli, and Alan~C Bovik.
\newblock Multiscale structural similarity for image quality assessment.
\newblock In {\em Asilomar Conference on Signals, Systems \& Computers}, pages
  1398--1402, 2003.

\bibitem{wiles2018x2face}
Olivia Wiles, A Koepke, and Andrew Zisserman.
\newblock X2face: A network for controlling face generation using images,
  audio, and pose codes.
\newblock In {\em European Conference on Computer Vision (ECCV)}, pages
  670--686, 2018.

\bibitem{wu2018video}
Chao-Yuan Wu, Nayan Singhal, and Philipp Krahenbuhl.
\newblock Video compression through image interpolation.
\newblock In {\em European Conference on Computer Vision (ECCV)}, pages
  416--431, 2018.

\bibitem{zaheer2017deep}
Manzil Zaheer, Satwik Kottur, Siamak Ravanbakhsh, Barnabas Poczos, Russ~R
  Salakhutdinov, and Alexander~J Smola.
\newblock Deep sets.
\newblock In {\em Neural Information Processing Systems (NeurIPS)}, volume~30,
  2017.

\bibitem{zakharov2020fast}
Egor Zakharov, Aleksei Ivakhnenko, Aliaksandra Shysheya, and Victor Lempitsky.
\newblock Fast bi-layer neural synthesis of one-shot realistic head avatars.
\newblock In {\em European Conference on Computer Vision (ECCV)}, pages
  524--540, 2020.

\bibitem{zakharov2019few}
Egor Zakharov, Aliaksandra Shysheya, Egor Burkov, and Victor Lempitsky.
\newblock Few-shot adversarial learning of realistic neural talking head
  models.
\newblock In {\em IEEE International Conference on Computer Vision (ICCV)},
  pages 9459--9468, 2019.

\bibitem{zhang2018unreasonable}
Richard Zhang, Phillip Isola, Alexei~A Efros, Eli Shechtman, and Oliver Wang.
\newblock The unreasonable effectiveness of deep features as a perceptual
  metric.
\newblock In {\em IEEE Conference on Computer Vision and Pattern Recognition
  (CVPR)}, pages 586--595, 2018.

\bibitem{zollhofer2018state}
Michael Zollh{\"o}fer, Justus Thies, Pablo Garrido, Derek Bradley, Thabo
  Beeler, Patrick P{\'e}rez, Marc Stamminger, Matthias Nie{\ss}ner, and
  Christian Theobalt.
\newblock State of the art on monocular 3d face reconstruction, tracking, and
  applications.
\newblock In {\em Computer Graphics Forum}, volume~37, pages 523--550, 2018.

\end{thebibliography}
}

\FloatBarrier

\FloatBarrier

\subsection*{H264 compression details}
We use ffmpeg 4.4 and the following command to for H264 compression in our comparison experiments
\\

\noindent {\tt ffmpeg -i \$INPUT\_FILE -preset medium -codec:v libx264 -x264-params bframes=0 -pix\_fmt yuv420p -na -crf \$CRF \$OUTPUT\_FILE}
\\

\noindent where {\tt \$CRF} is the quality factor.

\end{document}